\documentclass{article}

\PassOptionsToPackage{numbers, compress}{natbib}

\usepackage[preprint]{neurips_2023}

\usepackage[utf8]{inputenc} 
\usepackage[T1]{fontenc}    
\usepackage{hyperref}       
\usepackage{url}            
\usepackage{booktabs}       
\usepackage{amsfonts}       
\usepackage{nicefrac}       
\usepackage{microtype}      
\usepackage{xcolor}         
\usepackage{amsmath}
\usepackage{algorithmic}
\usepackage{algorithm}
\usepackage[pdftex]{graphicx}
\usepackage{paralist}

\usepackage[font=small,labelfont=bf]{caption}

\newcommand{\comm}[1]{}
\usepackage{color}
\newcommand{\Real}[0]{\mathbb{R}}

\newcommand{\michalis}[1]{{\color{blue}[Michalis: #1]}}

\title{Kalman Filter for Online Classification \\ of Non-Stationary Data}

\author{
    Michalis K. Titsias\thanks{Joint first authorship} \\
    Google DeepMind \\
    \texttt{mtitsias@google.com} \\
    \And
    Alexandre Galashov\footnotemark[1] \\
    Google DeepMind \\
    \texttt{agalashov@google.com} \\
    \And
    Amal Rannen-Triki \\
    Google DeepMind \\
    \texttt{arannen@google.com} \\
    \And
    Razvan Pascanu \\
    Google DeepMind \\
    \texttt{razp@google.com} \\
    \And
    Yee Whye Teh \\
    Google DeepMind \\
    \texttt{ywteh@google.com} \\
    \And
    Jörg Bornschein \\
    Google DeepMind \\
    \texttt{bornschein@google.com}
}

\begin{document}

\maketitle

\begin{abstract}
In Online Continual Learning (OCL) a learning system receives a stream of data and sequentially performs prediction and training steps. Important challenges in OCL are concerned with automatic adaptation to the particular non-stationary structure of the data, and with quantification of predictive uncertainty. Motivated by these challenges we introduce a probabilistic Bayesian online learning model by using a (possibly pretrained) neural representation and a state space model over the  linear predictor weights. Non-stationarity over the linear predictor weights is modelled using a “parameter drift” transition density, parametrized by a coefficient that quantifies forgetting. 
Inference in the model is implemented with efficient Kalman filter recursions which track the posterior distribution over the linear weights, while online SGD updates over the transition dynamics coefficient allows to adapt to the non-stationarity seen in data. While the framework is developed assuming a linear Gaussian model, we also extend it to deal with classification problems and for fine-tuning the deep learning representation. In a set of experiments in multi-class classification using data sets such as CIFAR-100 and CLOC we demonstrate the predictive ability of the model and its flexibility to capture non-stationarity. 
%
\end{abstract}

\section{Introduction \label{sec:intro}}
 
Continual Learning~\citep[e.g.][]{HADSELL20201028,parisi2019continual} is an open problem that has been receiving 
increasing 
attention in recent years. At its core, it aims to provide answers on how to train and use models in non-stationary scenarios. 
A multitude of different and sometimes conflicting desiderata have been
considered for continual learning, including forward transfer, backward transfer, avoiding catastrophic forgetting and
maintaining plasticity~\citep{HADSELL20201028}. Training and evaluation protocols highlight different 
constraints such as limited memory to store examples, limited model size or computational constraints.

In this work we focus on the Online Learning (OL)~\cite{ShalevSchwarzOL, Hazan16} scenario, where a learner receives a sequence of inputs $x_n$ and prediction targets $y_n$. At each time-step $n$, the model first observes $x_n$, makes a prediction and then receives the associated loss and ground truth target for learning. Within the deep learning community this scenario has also been referred to as \emph{Online Continual Learning} (OCL)~\citep{cai2021online, ghunaim2023real}, with the focus shifted more towards obtaining the best empirical performance on some given data instead of bounding the worst-case regret. 
%
Note that in OL/OCL each observation $(x_n, y_n)$ is first used for evaluation before it is used for training. 
No separate evaluation-sets are required and this objective naturally supports task-agnostic and 
non-stationary scenarios. Depending on the nature of the data stream, plasticity, not-forgetting and 
sample-efficiency all play a crucial role during learning. 
Furthermore, when considering the cumulative next-step log-loss under this protocol, it directly 
corresponds to the prequential description length and is thus a theoretically well motivated evaluation 
metric for non-stationary scenarios under the Minimum Description Length principle \citep{Grunwald2004, Blier2018-rl, bornschein2022sequential}.

In this paper, we propose a new method based on Kalman filters which explicitly takes into account non-stationaries in the data stream. 
It assumes a prior Markov model over the linear predictor weights that uses a “parameter drift”  transition density parametrized by a coefficient that quantifies forgetting. The prior model is then combined with observations using online Bayesian updates, implemented by computationally fast Kalman filter recursions, that track the posterior distribution over the weights of the linear predictor as the data distribution changes over time. 
These Bayesian updates are also combined with online SGD updates over the forgetting coefficient, thus allowing for more flexible  adaptation to non-stationarity. While the theoretical framework is developed assuming the tractable linear Gaussian case, we also extend the method to classification and for updating deep learning representations. 

We follow the trend introduced by large scale pretrained models, sometimes referred as foundation models~\citep{Bommasani2021OnTO}, of separating the representation learning from the readout classifier. These foundation models are seen as providing generic representations that can be used across a multitude of tasks, and in continual learning 
have been argued to provide stable representations, moving the focus of dealing with the non-stationarity to the convex readout layer. 
This separation of concerns allows a more natural integration of non-neural solution to deal with the continual or online problem. Following this perspective, we explore the performance of Kalman filter on top of a frozen pretrained representation. Furthermore, we also demonstrate that simultaneously online learning the representation and performing Kalman filter updates leads to stable learning and provide strong results on online continual learning benchmarks.

\section{Online learning with Kalman filter \label{sec_method}}

We first describe the method in an univariate regression setting and discuss in Section \ref{sec:classification} how we  adapt it to classification.
We assume a stream of data arriving sequentially  so that at step  $n$ we receive  $(x_n, y_n)$ where $x_n \in \Real^d$ is the input vector 
and $y_n \in \Real$ is the output. 
Our objective is to introduce an online learning model that captures non-stationarities and learns 
a good predictor for future data.        
Our model consists of two parts. The fist part is a deep neural network that outputs    
a representation $\phi(x;\theta) \in \Real^m$ so that $\phi_n := \phi(x_n;\theta)$ denotes the feature vector of the $n$-th data point. 
$\theta$ is a set of parameters that could be fixed, if the feature extractor  is a \emph{pretrained} network, 
or learnable, when it is either fine-tuned or learnt from scratch. 
Given $\phi_n$  the output $y_n$  is modelled by a Gaussian likelihood 
\begin{equation}
p(y_n | w_n) = \mathcal{N}(y_n | w_n^\top \phi_n, \sigma^2),
\label{eq:gaussianlikelihood}
\end{equation}
where the $m$-dimensional vector of 
regression coefficients $w_n$ depends on time index $n$.  
In other words, $w_n$ can change with time in order to capture distributional changes that may occur. 

The second part of model is an explicit prior assumption about how $w_n$ can change over time. More precisely, we will use a simple Gaussian Markov model over the parameters $w_n$, and model \emph{parameter transition or drift} by the
 following process,
\begin{align} 
\label{eq:w1}
p(w_0) & =\mathcal{N}(w_0 |0, \sigma_w^2 I), \ \ \ \ \  \ \ \ \ \ \ \ \ \ \ \ \ \ \ \ \ \ \ \ \ \ \ \ \ \ \ \ \ \ \ \ \ \ \ \ \ \ \ \text{Initial parameter}  \\ 
p(w_n|w_{n-1}) & = \mathcal{N}(w_n | \gamma_n w_{n-1}, (1 - \gamma_n^2) \sigma_w^2 I), \ \ n\geq1.  \ \ \ \ \text{Parameter drift}
\label{eq:wn_wnm1}
\end{align}
The time-dependent parameter $\gamma_n$ takes values in $[0, 1]$ and quantifies the \emph{memory or forgetting} of the process. 
For example, 
when  $\gamma_n=1$  then $w_n=w_{n-1}$, which means that at time step $n$ the model re-uses (or copies forward) the parameter $w_{n-1}$ from the  
previous step. Such extreme case is suitable when there is no distributional change at time $n$. In the other extreme case, when $\gamma_n=0$ 
the parameter $w_n$ is fully refreshed, i.e.\ reset to the 
prior $\mathcal{N}(0, \sigma_w^2 I)$, which implies a sharp change in the distribution. Similarly, intermediate values of $\gamma_n \in (0,1)$ can model more smooth 
or gradual changes. Flexible learning of $\gamma_n$ through time  will be a key element of our method that we describe in Section \ref{sec:learninggammaclass} for the  classification problem. Furthermore,  it is worth noting that the transition model in \eqref{eq:wn_wnm1} 
is a variance-preserving diffusion, 
so that when $w_{n-1} \sim \mathcal{N}(0, \sigma_w^2 I)$  
the next state $w_n = \gamma_n w_{n-1} + \sqrt{1-\gamma_n^2} \epsilon$ ($\epsilon \sim \mathcal{N}(0, I)$) follows also $\mathcal{N}(0, \sigma_w^2 I)$.  Hence, the variance $\sigma_w^2$ of the regression parameters remains constant  through time.           
 
Having specified the observation model in \eqref{eq:gaussianlikelihood} and the transition model over the parameters $w_n$ in \eqref{eq:w1}-\eqref{eq:wn_wnm1} 
we can write the full joint density up to the $n$-th observation as 
\begin{equation} 
p(w_0) \prod_{i=1}^n p(y_i | w_i) p(w_i | w_{i-1})  = 
\mathcal{N}(w_0|0,\sigma_w^2 I) \prod_{i=1}^n  \mathcal{N}(y_i | w_i^\top \phi_i, \sigma^2) \mathcal{N}(w_i | \gamma_i w_{i-1},  (1 - \gamma_i^2) \sigma_w^2 I).
\label{eq:joint} \nonumber
\end{equation}
 %
%
Clearly, this is a type of a linear Gaussian state space model where exact online Bayesian inference over $w_n$ can be solved with standard recursions.    
Specifically, inference can be carried out by standard Kalman filter prediction and update steps \citep{sarkka_2013} to compute online the Bayesian posterior over $w_n$. The prediction step 
requires computing the posterior over $w_n$ given all past data up to time  $n-1$ (and excluding the current $n$-th observation),  which is a Gaussian 
density $p(w_n | y_{1:n-1}) 
= \mathcal{N}(w_n | m_n^{-}, A_n^{-})$
with parameters
\begin{equation}
m_n^{-}  = \gamma_n m_{n-1},  \ \ \ 
A_n^{-}  = \gamma_n^2 A_{n-1}  +  (1-\gamma_n^2) \sigma_w^2 I.
\label{eq:m_A_minus}
\end{equation}
The update step finds the updated Gaussian posterior $p(w_n | y_{1:n}) = \mathcal{N}(w_n | m_n, A_n)$ by modifying the mean vector $m_n$ and the covariance matrix
$A_n$  to incorporate the information coming from the most recent 
observation $(x_n,y_n)$ according to   
\begin{equation}
m_n  = m_n^- 
+ \frac{A_n^- \phi_n}{\sigma^2 + \phi_n^\top A_n^- \phi_n}
(y_n - \phi_n^\top m_n^-), \ \ \
A_n  = A_n^-
- \frac{A_n^- \phi_n \phi_n^\top A_n^-}{\sigma^2 + \phi_n^\top A_n^- \phi_n}.
\label{eq:kalmanstep}
\end{equation}
The initial conditions  for the recursions are $m_0 = 0$ and $A_0 = \sigma_w^2 I$. 
The cost per iteration is 
$O(m^2)$ which means that the Kalman recursions are computationally very efficient.     

Further, if we remove the stochasticity from the transition dynamics by setting $\gamma_n=1$ for any $n$, 
the above Kalman recursions reduces to online Bayesian linear regression as detailed in Appendix \ref{app:connectionBLR}.    

\paragraph{Prediction.} As the online model sequentially receives observations and updates its Bayesian posterior distribution over $w_n$, 
it can also perform next step predictions. Suppose that after $n-1$ steps the model has observed the data $(x_i, y_i)_{i=1}^{n-1}$ 
and computed the posterior $p(w_n | y_{1:n-1})$.  Then for the  next input data $x_n$ the model 
can predict its output $y_n$   based on the Bayesian predictive density 
\begin{equation}
p(y_n | y_{1:n-1}) =
\int p(y_n|w_n) p(w_n | y_{1:n-1}) d w_n  =  \mathcal{N}(y_n | \phi_n^\top m_n^-, \phi_n^\top A_n^{-}  \phi_n + \sigma^2),
\label{eq:gaussianpredictive}
\end{equation}
which is analytic for this Gaussian regression case, while for the classification case will require approximate inference and Monte Carlo sampling 
as detailed in Section \ref{sec:learninggammaclass}.

\subsection{Application to classification
\label{sec:classification}
}

We now adopt the above online learning model to multi-class classification problems. 
Suppose a classification problem with $K$ classes,  where the
label $y_n$ is encoded as $K$-dimensional one-hot vector, i.e.\  $y_n \in \{0,1\}^K$, $\sum_{k=1}^K y_{n,k}=1$.  
A suitable observation model for classification is the standard softmax likelihood 
$
p(y_{n,k}= 1 | W_n) = \frac{\exp\{w_{n,k}^\top \phi_n\}}{\sum_{j=1}^K \exp\{w_{n,j}^\top \phi_n\} },  
$
where $W_n = (w_{n,1}, \ldots, w_{n,K})$ is a $m \times K$ matrix storing  the parameters $W_n$ which play a similar role to previous 
regression coefficients. These parameters follow $K$ independent Markov processes so that each $k$-th   
column $w_{n,k}$ of $W_n$  is independent from the other columns and 
obeys the transition dynamics from \eqref{eq:w1}-\eqref{eq:wn_wnm1}.  However, unlike the regression
case, exact online inference over the parameters $W_n$ 
using Kalman recursions is intractable due 
to the non-Gaussian form of the softmax likelihood. Thus, we need to rely on approximate inference. 
Next we derive a fast and very easy to implement inference technique that still uses the exact  
Kalman recursions. This consists of two main components described next.    


 \paragraph{(i) Maintain a fast and analytic Kalman recursion.} To achieve this we introduce a Gaussian likelihood that (as a means to approximate inference) replaces the softmax likelihood. It
has the form
$
q(y_n|W_n) = \prod_{j=1}^K \mathcal{N}(y_{n,j} | w_{n,j}^\top \phi_n, \sigma^2),
$
which explains the elements of the one-hot vector 
by a Gaussian density, a trick that has been used successfully in the literature e.g.\ for Gaussian process classification \citep{rasmussen2006gaussian} and meta learning \citep{patacchiola2019deep}.  
With this approximate likelihood the Kalman recursions remain tractable and propagate forward 
an approximate predictive posterior $q(W_n|y_{1:n-1}) = \prod_{k=1}^K \mathcal{N}(w_{n,k}| 
m_{n,k}^-, A_n^-)$  with mean parameters  given by the $m \times K$ matrix $M_n^-=(m_{n,1}^-, \ldots, m_{n,K}^-)$ 
and covariance parameters given by the $m \times m$ matrix $A_n^-$: 
\begin{equation}
\label{eq:means_minus}
M_n^{-}  = \gamma_n M_{n-1}, \ \ \ 
A_n^{-} = \gamma_n^2 A_{n-1}  + (1 -\gamma_n^2) \sigma^2_w I. 
\end{equation}
The corresponding updated posterior   
 $q(W_n|y_{1:n}) = \prod_{k=1}^K \mathcal{N}(w_{n,k}| 
m_{n,k}, A_n)$ 
has parameters 
\begin{equation}
\label{eq:means_plus}
M_n  = M_n^- 
+ \frac{A_n^- \phi_n}{\sigma^2 + \phi_n^\top A_n^- \phi_n} \times 
(y_n^\top - \phi_n^\top M_n^-), \ \ \ 
A_n  = A_n^-
- \frac{A_n^- \phi_n \phi_n^\top A_n^-}{\sigma^2 + \phi_n^\top A_n^- \phi_n}.
\end{equation}
The recursion is initialized at  $M_0 = {\bf 0}, A_0 =\sigma^2_w I$, where ${\bf 0}$ denotes the $m \times K$ matrix of zeros. A full iteration costs $O(K m + m^2)$ and if the size $m$ of the
feature vector is larger than the number of classes $K$, the term $O(m^2)$ dominates and the complexity is the same as in univariate regression. 
The crucial factor to obtain such an efficiency is that the covariances matrices $(A_n^-,A_n)$ are shared among all $K$ classes, which is because 
the hyperparameter $\sigma^2$ in the approximate Gaussian likelihood 
is shared among all $K$ dimensions.\footnote{If we
choose a different $\sigma_k^2$ per class the time and storage complexity grows to $O(k m^2)$ which is too expensive in practice.} 

\paragraph{(ii) Kalman posteriors interact with softmax for prediction or parameter updating.} We
 view the above Kalman recursion 
 as an online approximate inference procedure that provides an approximation to the 
 exact  
posterior distribution. For example, if  $p(W_n|y_{1:n-1})$ is the exact intractable predictive posterior (obtained by Bayes' rule with the exact softmax likelihood) then 
 $q(W_n|y_{1:n-1})$ computed from Kalman recursion is an approximation to $p(W_n|y_{1:n-1})$. 
 Subsequently, by following standard approximate Bayesian inference practices, whenever we wish to predict class probabilities or compute a cross entropy-like loss 
to optimize parameters, the approximate posterior will be combined with the exact softmax through Bayesian 
averaging and Monte Carlo estimation. Below we make use of this in two cases, one for online learning the forgetting coefficient $\gamma_n$ discussed
in Section \ref{sec:learninggammaclass}, and the second  in Section \ref{sec:finetuning}
for computing accurate predictive probabilities and fine-tuning the representation. For the second case we also find it 
useful to introduce a Bayesian calibration procedure, which optimizes online a calibration parameter and improves  drastically 
the predictive probability estimates.   
      
 \subsubsection{Online updating the forgetting coefficient $\gamma_n$
\label{sec:learninggammaclass}   
}

An important aspect of our method is the online updating of the forgetting coefficient $\gamma_n$, which is 
indexed by $n$ to indicate that it can change over time to reflect the distributional changes seen in the data stream. 
Instead of using a full Bayesian approach (that will require an extra hierarchical prior assumption) over $\gamma_n$ 
we will follow a simple empirical Bayes method where we update $\gamma_n$ by online point estimation.   
  For that, we first initialize $\gamma_0$  and then for any subsequent 
time step $n\geq1$, where we observe $(x_n, y_n)$, we first copy the previous $\gamma_{n-1}$ value to the new step, i.e.\ $\gamma_n = \gamma_{n-1}$, and then  
we apply an SGD update to change $\gamma_n$ by maximizing the log predictive probability. 
Schematically, the SGD update is written as  
\begin{equation}
\gamma_n \leftarrow 
\gamma_n + \rho_n \nabla_{\gamma_n} \log p(y_{n,k}=1 | y_{1:n-1}),
\label{eq:updategamma}
\end{equation}
where we further parametrize 
$\gamma_n = \exp( - 0.5 \delta_n)$, with $\delta_n \geq 0$, 
so that the update is applied to $\delta_n$.\footnote{The hard constraint $\delta_n \geq 0$ is imposed through clipping.}    
However,  since the log predictive probability is not tractable 
we consider the  approximation 
\begin{equation}
\log p(y_{n,k}=1 |y_{1:n-1}) \approx
 \log \int p(y_{n,k}=1  |W_n) q(W_n|y_{1:n-1}) d W_n, 
 \label{eq:eq:classification_log_pred1}
\end{equation}
where $p(y_n|W_n)$ is the softmax 
and 
$
q(W_n|y_{1:n-1}) 
= \prod_{k=1}^K \mathcal{N}(w_{n,k}| 
m_{n,k}^-, A_n^-)$ is the Kalman analytic approximate posterior distribution. To estimate this we can use the standard procedure to reparametrize 
the integral in terms of the $K$-dimensional vector of logits 
 $f_n = W_n^\top \phi_n$ which follows the factorized Gaussian distribution $q(f_n | \mu_n,  s_n^2 I)$ with mean    
$\mu_n = (M_n^-)^\top \phi_n$ and isotropic variance  $s_n^2 = \phi_n^\top A_n^- \phi_n$.  After this reparametrization,  
 \eqref{eq:eq:classification_log_pred1} leads to the estimate  
\begin{align}
 \label{eq:classification_log_pred2}
\log p(y_{n,k}=1 |y_{1:n-1})  & = \log \int \frac{\exp\{f_{n,k}\}}{\sum_{j=1}^K \exp\{ f_{n,j}\} }
q(f_n | \mu_n,  s_n^2 I) d f_n 
\\ 
\label{eq:montecarlo_log_pred} 
&  \approx \log \frac{1}{S} 
 \sum_{s=1}^S 
  \frac{\exp\{\mu_{n,k} +  s_n \epsilon^{(s)}_k \}}{\sum_{j=1}^K \exp\{ \mu_{n,j}  + s_n \epsilon^{(s)}_j \} },  \ \  \epsilon^{(s)} \sim \mathcal{N}(0,I), 
\end{align}
where to move from \eqref{eq:classification_log_pred2} 
to (12) 
we first reparametrize the integral to be an expectation under the standard normal and then apply Monte Carlo. 
To update $\gamma_n$ using SGD, we differentiate this final Monte Carlo estimate of the loss $- \log p(y_{n,k}=1 |y_{1:n-1})$ wrt the parameter $\delta_n$  in $\gamma_n = \exp\{-0.5 \delta_n \}$, where $\gamma_n$  appears in
 $(\mu_n, s_n^2)$ through the computation  $M_n^{-} = \gamma_n M_{n-1},  A_n^{-} = \gamma_n^2 A_{n-1}  + \sigma_w^2 (1 -\gamma_n^2) I$, while the parameters $(M_{n-1}, A_{n-1})$ of the posterior up to time $n-1$ 
are taken as constants.

\subsubsection{Calibration of class probabilities and fine-tuning the representation 
\label{sec:finetuning}
}


Given that in classification the predictive posterior $q(W_n|y_{1:n-1})$ is an (possibly crude) 
approximation to the true posterior, the log predictive probability estimate in (12) 
 can be inaccurate. We improve this estimate
by fine-tuning over a \emph{calibration parameter} that is optimized with gradient steps.  In the same step we can also fine-tune the neural network parameters $\theta$  that determine the representation vector $\phi(x,\theta)$. 
Specifically, for the calibration  procedure we introduce a parameter $\alpha > 0$ that rescales the logits inside the softmax function, so that 
the softmax in  \eqref{eq:classification_log_pred2}  is replaced by  $\frac{\exp\{\alpha f_{n,k}\}}{\sum_{j=1}^K \exp\{ \alpha f_{n,j}\} }$ and the final
 Monte Carlo estimate becomes \begin{align} 
 \log p(y_{n,k}=1 |y_{1:n-1}, \alpha, \theta)  \approx \log \frac{1}{S} 
 \sum_{s=1}^S 
  \frac{\exp\{\alpha \mu_{n,k} +  \alpha s_n \epsilon^{(s)}_k \}}{\sum_{j=1}^K \exp\{ \alpha \mu_{n,j}  + \alpha s_n \epsilon^{(s)}_j \} }.
  \label{eq:calibratedlogprob}
  \end{align}  
  Then the negative log predictive probability, 
  $ - \log p(y_{n,k}=1 |y_{1:n-1}, \alpha, \theta)$, is treated as a loss that is optimized jointly over $(\alpha,\theta)$  with online SGD steps, i.e.\ as individual
  data points $(x_n, y_n)$ arrive sequentially. 
This resembles the forward-calibration described in \citep{bornschein2022sequential}.
Algorithm \ref{alg:kalman_algorithm} summarizes the whole online learning procedure that includes the Kalman filter recursion,  update of coefficient
$\gamma_n$ and fine-tuning of the representation parameters $\theta$ and calibration parameter $\alpha$. An ablation study in Appendix \ref{app:calibration} shows that the calibration procedure can significantly improve the log predictive probability estimates.    

  
The fully online process for fine-tuning  the deep network parameters $\theta$ in Algorithm  \ref{alg:kalman_algorithm}  can be expensive since modern hardware 
is computationally more effective when forward and backpropagation passes in deep neural nets are applied jointly to minibatches rather to individual data points. In practice, we therefore also consider a faster version of Algorithm \ref{alg:kalman_algorithm} where we predict ahead a batch of $b$ data points by treating them as i.i.d.\ and then take a gradient step over $(\alpha, \theta)$, i.e.\ we use as 
loss  $-\frac{1}{b} \sum_{i=1}^b \log p( y_{n+i} | y_{1:n-1}, \alpha, \theta), i=1, \ldots, b$. This creates two options for the Markov dynamics: (i) either apply the transition "batch-wise" where the transition is taken every $b$ data forming the minibatch (so that time index $n$ corresponds to the number of minibatches), or (ii) the minibatch updating only affects the prediction over future data (we predict ahead $b$ points instead of only the next one) and the fine-tuning of $(\alpha, \theta)$,  
while the subsequent Kalman recursion steps and SGD update over $\gamma_n$ are applied online, i.e.\ by processing the $b$ data in the minibatch one by one. 
Both schemes can be useful in practice, as further discussed in the Appendix. For simplicity, Algorithm  \ref{alg:kalman_algorithm}
presents the purely online version, while pseudocode for the above minibatch-based variants is similar. 


\begin{algorithm}[tb]
   \caption{Kalman filter online learning}
   \label{alg:kalman_algorithm}
\begin{algorithmic}
   \STATE {\bfseries Input:} Data stream $(x_n,y_n)_{n \geq 1}$; representation $\phi(x, \theta)$; parameters $\theta$; 
   hyperparameters $(\sigma^2, \sigma_w^2)$ with default values $(1/K, 1/m)$.   
   \STATE Initialise $M_0 = {\bf 0}, A_0 =\sigma_w^2 I$ and $\delta_0$ (e.g. to value 0.0 so that $\gamma_0=1$).
   \FOR{data point $n=1,2,3,\ldots,$}
   \STATE Observe $x_n$, compute $\phi_n = \phi(x_n, \theta)$ make a prediction for the label $y_n^*$ via eq. \eqref{eq:calibratedlogprob}
   \STATE Observe true class label $y_n$
   \STATE (Optional) Fine-tune the deep neural network parameters $\theta$ together with the calibration parameter $\alpha$, as described in Section \ref{sec:finetuning}   
   \STATE Update $\delta_n$ as described in Section  \ref{sec:learninggammaclass} and set $\gamma_n = \exp\{-0.5\delta_n\}$
     \STATE 
     $M_n^{-}  = \gamma_n M_{n-1}, \  A_n^{-} = \gamma_n^2 A_{n-1}  + (1 -\gamma_n^2) \sigma^2_w I$  
   \STATE Update Kalman statistics:
   $M_n  = M_n^- + \frac{A_n^- \phi_n}{\sigma^2 + \phi_n^\top A_n^- \phi_n} \times  (y_n^\top - \phi_n^\top M_n^-), \  A_n  = A_n^- - \frac{A_n^- \phi_n \phi_n^\top A_n^-}{\sigma^2 + \phi_n^\top A_n^- \phi_n}$  
   \ENDFOR
\end{algorithmic}
\end{algorithm}

\section{Related Work \label{sec_related}} 

\paragraph{Continual learning to reduce forgetting.}
Classical approaches for continual learning have been developed with a focus on reducing forgetting~\citep{de2021continual, parisi2019continual, mai2022online}. Most of these approaches could be separated into three different families: \emph{Replay or sampling based approaches} include~\citep{chaudhry2019continual, aljundi2019gradient, sun2022information, aljundi2019online, jin2020gradient, lopez2017gradient, shin2017continual, wang2022learning}, \emph{Regularization based approaches} include~\citep{kirkpatrick2017overcoming, zenke2017continual,aljundi2018memory, li2017learning, zhang2020class, buzzega2020dark, caccia2022new} and \emph{Parameter isolation approaches} include~\citep{mallya2018packnet, rusu2016progressive, saha2021gradient}. Other approaches have also been proposed, such as
~\cite{lee2020neural} where an expansion-based approach is proposed, leveraging a set of experts trained and expanded using a Bayesian non-parametric framework. More recently, the trend moved the focus towards forward transfer and efficient adaptation~\citep{hadsell2020embracing, bornschein2022nevis, ghunaim2023real, hayes2022online}.

While a significant progress has been achieved through this line of research, one major drawback, highlighted by different recent works~\citep{ghunaim2023real, cai2021online, caccia2022new}, is the focus on small-scale artificial benchmarks with abrupt and unnatural distribution shift, and on metrics that fail to capture the capability of the models to efficiently adapt to the non-stationarity in the input data. In this work, we focus on fast adaptation and the next-step prediction problem from OL/OCL.  

\paragraph{Online continual learning and fast adaptation.}
Different recent works considered the problem of fast adaptation and the set of constraints and objectives that are realistic in an online scenario. Different recent benchmarks, e.g. CLOC ~\citep{cai2021online} and CLEAR~\citep{lin2021clear}, propose a sequence of temporally sorted images with naturally shifting visual concepts. CLOC leverages a subset of 39M images from YFCC100M~\citep{thomee2016yfcc100m} along with their timestamps and geolocalisation tags, spanning a period of 9 years. The paper also highlights the limits of the classical continual learning approaches, and proposes simple baselines to overcome them, based mostly on adapting the online learning rate and the replay buffer size. They focus on an online evaluation protocol where each mini-batch is used for testing before adding it to the training dataset. 
In this work, we largely follow the CLOC setting. We evaluate our approach on the CLOC data and on modified CIFAR100 versions that are inspired by it. 

In \cite{bornschein2022sequential} the authors highlight the theoretical and practical benefits of using 
the sequential (online) next-step performance for model evaluation. They emphasize the connection to 
compression based inference and to the prequential Minimum Description Length principle~\citep{dawid1999prequential, poland2005asymptotics}. 
Empirically, they evaluate SGD based techniques on the pareto-front of prediction loss vs. computational requirements; 
and propose forward-calibration and specific rehearsal approaches to improve results. We compare against their methods in our experiments on CLOC. 


The authors of \cite{ghunaim2023real} base their work on the observation that CL approaches have been developed under unrealistic constraints, allowing for offline learning on data with multiple passes without any limitation of computational cost. In realistic settings where we aim at updating a model on continuously incoming data, approaches that are slower than the stream would be impractical, or reach suboptimal performance in the limited time and computational budget. The authors propose an evaluation protocol that puts the computational cost at the center. They propose adding a delay between model update and evaluation. This delay is related to the computational cost of the updating method. Under the same budget, a twice more expensive approach would update the model half as often. With this realistic evaluation, the authors show that a simple baseline based on experience replay outperforms state-of-the-art CL methods, due to their complexity. In this work, we follow the protocol described in \cite{ghunaim2023real}. Our approach has a relative complexity that requires no delay between training and evaluation. We therefore compare against the baselines reported in \cite{ghunaim2023real} that have the same properties, namely, ER~\citep{chaudhry2019continual}, ER++~\citep{ghunaim2023real} and ACE~\citep{caccia2022new}.

\paragraph{Kalman filters.}
The idea of relying on Kalman filters to estimate the posterior distribution over the weights of a neural networks dates back to the 90s, with work such as \cite[e.g][]{LanceEKA88,Feldkamp1998,Puskorius94}. The focus in these works is to accelerate learning by incorporating second order information in the step size, and the setting typically considered is that of stationary learning. Kalman filter for dealing with online learning has been explored in the linear case for example in \cite{tsiamis2020online,Kozdoba19}. In contrast, in our work we focus on the ability of Kalman filter to help with the online continual learning problem by relying on pretrained representations.

\section{Experiments \label{sec:experiments}}

\subsection{Illustrative time series example} 
\label{sec:illustrative_example}

We first apply our method to artificial time series data. 
The task is to track a non-stationary data stream of scalar noisy observations $y_n$ without any conditioning input $x_n$. 
We further assume a very simple model 
where the feature 
is just an univariate constant value equal to unity, i.e.\  $\phi_n=1$
so that 
the observation likelihood simplifies as   
$p(y_n | w_n) =  \mathcal{N}(y_n | w_n, \sigma^2)$ and the parameter $w_n$, 
to be inferred through time, models the unknown expected value of $y_n$.  
 
\begin{figure}
\centering
\begin{tabular}{cc}
\includegraphics[scale=0.25]
{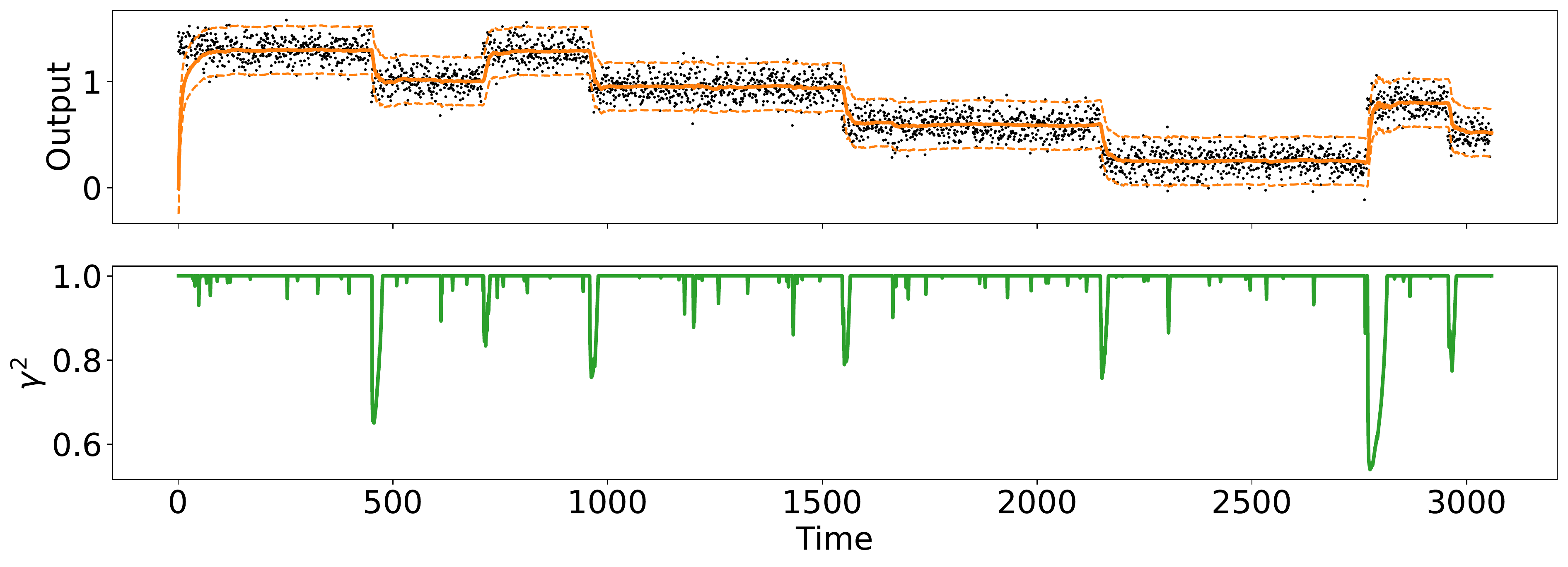} &
\includegraphics[scale=0.25]
{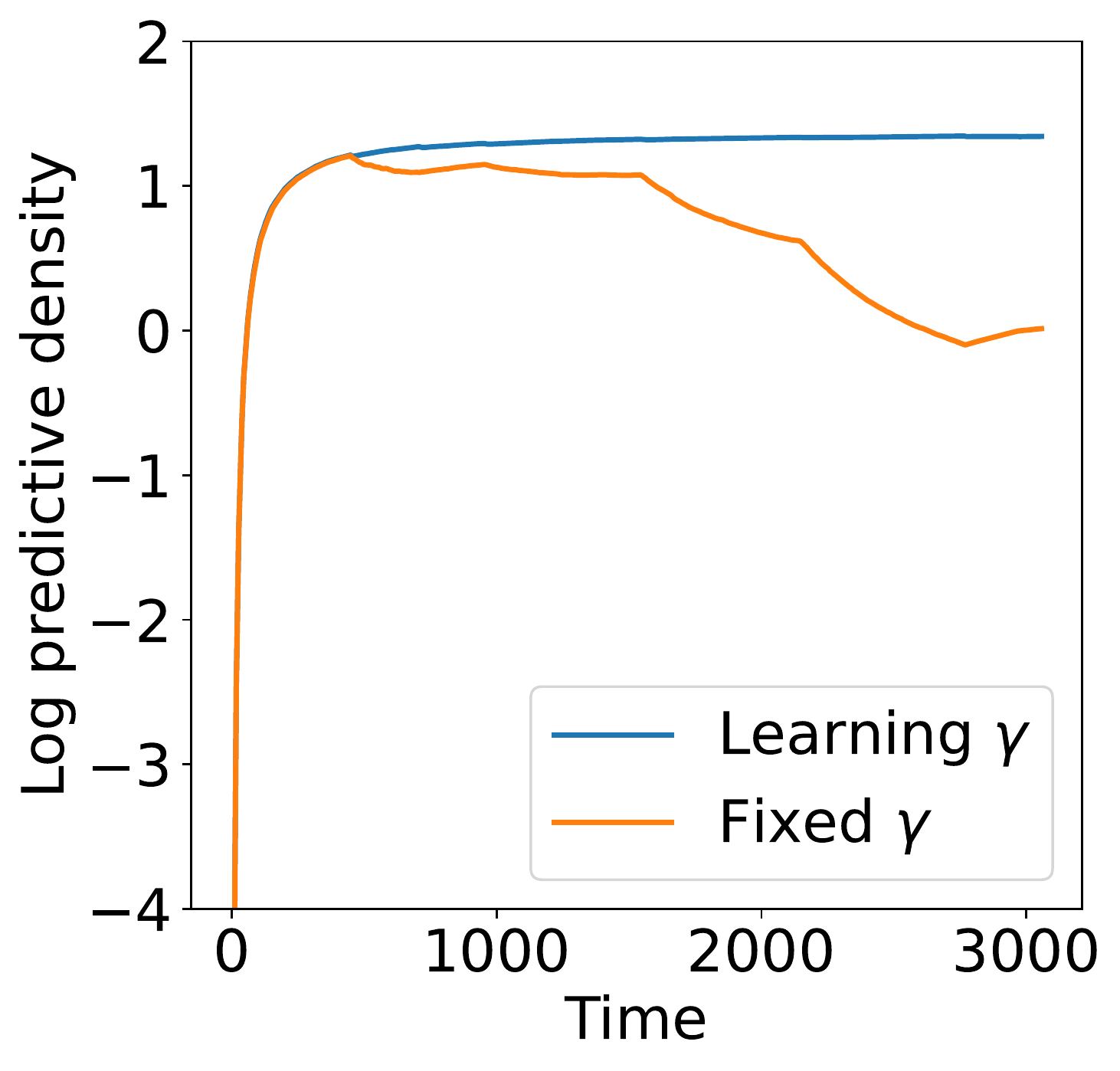} 
\end{tabular}
\caption{Results in a artificial time series example of $3058$ observations. Top row in left panel shows the data (black dots) 
and the predicted mean and uncertainty (orange lines) over $y_n$ (as data arrive sequentially from left to right and we perform online next step  prediction), while 
the bottom row shows the optimized values of $\gamma_n^2 = \exp(-\delta_n)$. Right panel shows the accumulated average log predictive density, i.e.\ 
$\frac{1}{n} \sum_{i=1}^n \log p(y_i | y_{1:i-1})$, computed across time for the model that learns $\gamma_n$ and the model that ignores non-stationarity by 
setting $\gamma_n=1$ for all $n$.        
} 
\label{fig:toyregression}
\end{figure}

Figure\ \ref{fig:toyregression} shows the results of the Kalman model that was initialized with $\gamma_0=1$ 
and learns it online, as described in Appendix \ref{app:learninggamma}. The non-stationary nature of the artificial series is such that the signal is
piece-wise (noisy) constant with seven change-points. As shown by  the second row in the left panel in Figure \ref{fig:toyregression} the learned value of $\gamma_n$ 
is able to adjust to this non-stationarity by dropping the value of $\gamma_n$ quite below the value one (in order to refresh the Bayesian statistics over $w_n$)
 any time there is a change-point. In contrast, if we remove the ability to capture  non-stationarity, i.e.\ by setting $\gamma_n=1$  for all $n$, 
 the performance gets much worse  as shown by the accumulated log predictive density scores in Figure\ \ref{fig:toyregression} and by Figure \ref{fig:toyregression_app} in the Appendix \ref{app:toyregression}.

\subsection{Online classification
\label{sec:classificationresults}
}

In this section, we study the behaviour of our method when applied to online classification. We consider the OCL
scenario where data arrives in small batches (chunks) \cite{cai2021online,bornschein2022sequential,ghunaim2023real}: The learning algorithm is exposed to the stream of data $\mathcal{S}$ such that every point of time $n$, a chunk of data $S_{n} = \{(x_{n,b + i}, y_{n, b+i})\}_{i=1}^b$ of size $b$ is revealed. 
The algorithm first predicts the labels $y_{s}, \; s=(nb+1),\ldots,(n+1)b$,
and then updates its parameters $\theta$ and $\alpha$ based on this chunk of data.
Like to before, we compute the \emph{Average Online Accuracy}:
\begin{equation}
    \label{eq:aoc}
    acc_o(n) = \frac{1}{n b} \sum_{s=1}^{n b} acc(y_{s}, \hat{y_s}),
\end{equation}
where $y_{s}$ is the ground truth label and $\hat{y}_s$ is the prediction of the model. This metric contains accuracies computed on the fly over the course of training and quantifies how well the learning algorithm ingests new knowledge.

\subsubsection{Online classification on CIFAR-100}

We evaluate the performance of the Kalman Filter on two variants of online classification on CIFAR-100~\citep{krizhevsky2009learning}: \emph{stationary online classifcation on CIFAR-100} and \emph{non-stationary online classifaction on CIFAR-100}. In the \emph{stationary case}, we follow the protocol described in \cite{ghunaim2023real} where the stream $S$ is constructed by randomly shuffling CIFAR-100 dataset and split into chunks, so that learning algorithm does one pass through CIFAR-100. This is similar to One-Pass ImageNet~\citep{hu2021pass} benchmark. Since it is randomly shuffled, there is no non-stationarity. In the \emph{non-stationary case}, we consider a task-agnostic class-incremental version of Split-CIFAR100 as in \citep{lee2020neural}, where CIFAR-100 is split into 10 tasks (task identity is not communicated to the learning algorithm) each containing 10 different classes, concatenated into a stream and split into chunks. At any time, the learning algorithm solves a multi-classification problem with $100$ classes. In this setting, there is very distinct non-stationarity related to the task changes. This benchmark is also studied by many class incremental continual learning approaches which focus on alleviating catastrophic forgetting~\citep{FRENCH1999128, MCCLOSKEY1989109} and using average \emph{incremental} accuracy as metric. Dealing with catastrophic forgetting goes beyond the scope of this work and we only focus on average \emph{online} accuracy. In both cases, we follow training/evaluation protocol described in \cite{ghunaim2023real} dealing with chunks of size $10$. For more details, see Appendix~\ref{app:exp_details}.

For Kalman filter variants we consider the following options. \emph{Stationary Kalman Filter ($\gamma=1$)}, which could be seen as a form of Bayesian Logistic Regression, \emph{Non-stationary Kalman Filter with fixed $\gamma=0.999$} and \emph{Non-stationary Kalman Filter with learned $\gamma$}. On top of that, we study performance of Kalman filter in three regimes: \emph{no backbone finetuning}, a regime with fixed randomly initialized features $\phi$ which is a linear model, \emph{backbone finetuning} and \emph{backbone finetuning with replay}. As baselines, we consider ER and ER++ from \cite{ghunaim2023real} as well as ACE~\citep{caccia2022new} (results are taken from \cite{ghunaim2023real}). The results are only reported for stationary CIFAR-100 since this was the only setting studied in \cite{ghunaim2023real}. The hyperparameters for methods are selected by choosing the variant with highest cumulative log probabilities following MDL principle from \citep{bornschein2022sequential}. See Appendix~\ref{app:exp_details} for more details.


The results are given in Table~\ref{table:online_cifar100}. In the \textbf{stationary} CIFAR-100 case, we observe that stationary Kalman filter provides a reasonable performance and is generally better than non-stationary Kalman filter with fixed $\gamma$. This is consistent with our intuition that there is not much non-stationarity to model and therefore we would not expect Kalman filter to help. We see that Kalman Filter with learning $\gamma$ leads to slightly better results than stationary case. Intuitively, it makes sense since in the worst case, this variant could revert back to $\gamma=1$. Moreover, we see that replay-free Kalman filter leads to very competitive results against external baselines. Adding replay to Kalman improves results even further, beating ER++ baseline which uses much more replay than ER (see \citep{ghunaim2023real}).

\begin{figure}
\centering
\includegraphics[scale=0.177]{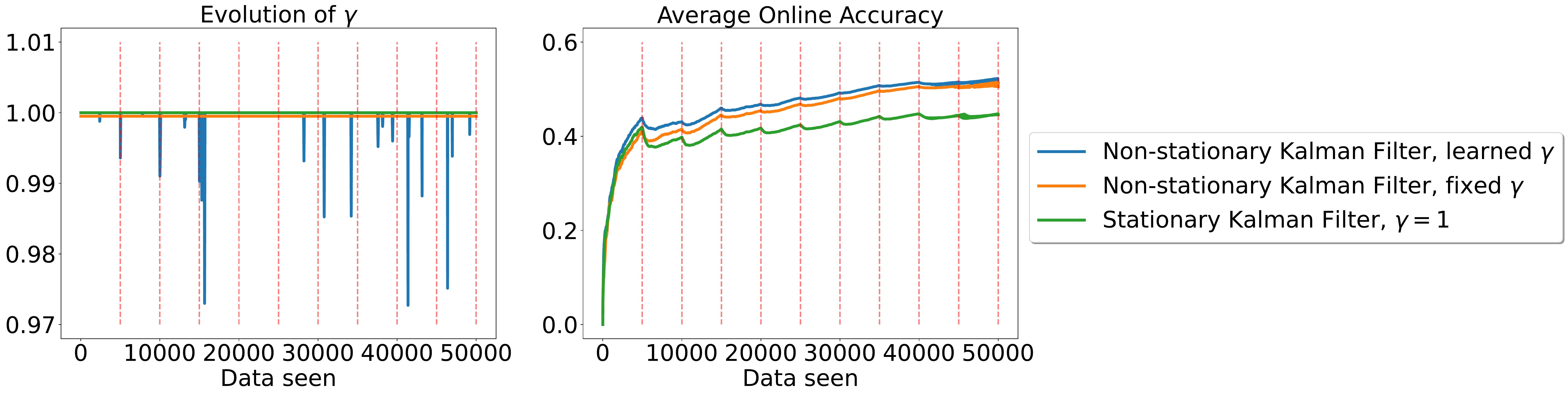}
\caption{\textbf{Non-stationary CIFAR-100}. The left plot shows the evolution of $\gamma$, the right plot shows the corresponding average online accuracy. The red dashed lines correspond to the task boundaries.} 
\label{fig:nonstationary_cifar100}
\end{figure}

In the \textbf{non-stationary} CIFAR-100 case, we see that stationary Kalman filter performs consistently worse than its non-stationary variants. This is consistent with our intuition since in this case, there is very specific non-stationarity which Kalman filter could capture. Moreover, we again see that learning $\gamma$ generally leads to better performance. Figure~\ref{fig:nonstationary_cifar100} visualizes the dynamics of learning $\gamma$ in case of \emph{Backbone Finetuning} setting, with red dashed lines indicating task boundaries. We see that in many cases, $\gamma$ drops at task boundaries, essentially pushing down probabilities of classes from previous classes and focusing more on future data. This is the desired behaviour of the method since it essentially captures non-stationarity.



\begin{table}[!htbp]
\small
\centering
\caption{\textbf{CIFAR-100} results in stationary and non-stationary settings. The numbers in \textbf{bold} correspond to the best performing method in the group. Results for external baselines are taken from \citep{ghunaim2023real}. The amount of replay used for Kalman filter is similar to ER baseline.} 
\label{table:online_cifar100}
\begin{tabular}{|l|c|c|}
\toprule
 & \multicolumn{2}{c|}{Average Online Accuracy} \\
\midrule
Method & Stationary CIFAR-100 & Non-stationary CIFAR-100 \\
\midrule
\textbf{\emph{No backbone finetuning (Purely linear model)}} & & \\
Stationary Kalman Filter ($\gamma=1.0$) & 10.9 \% & 12.1 \% \\
Non-stationary Kalman Filter (fixed $\gamma=0.999$) & 9.2 \% & 31.9 \% \\
Non-stationary Kalman Filter (learned $\gamma$) & \textbf{11.4 \%} & \textbf{32.7} \% \\
\midrule
\textbf{\emph{Backbone finetuning}} & & \\
Stationary Kalman Filter ($\gamma=1.0$) & 16.4 \% & 44.5 \% \\
Non-stationary Kalman Filter (fixed $\gamma=0.999$) & 15.9 \% & 50.5 \% \\
Non-stationary Kalman Filter (learned $\gamma$) & \textbf{16.9 \%} & \textbf{51.2 \%} \\
\midrule
\textbf{\emph{Backbone finetuning with Replay}} & & \\
Stationary Kalman Filter ($\gamma=1.0$) & 18.5 \% & 51.6 \% \\
Non-stationary Kalman Filter (fixed $\gamma=0.999$) & 18.9 \% & 55.5 \% \\
Non-stationary Kalman Filter (learned $\gamma$) & \textbf{19.0 \%} & \textbf{55.5 \%} \\
\midrule
\textbf{\emph{External baselines}} & & \\
ACE~\citep{caccia2022new} & 14.42\% & -- \\
ER & 13.62\% & -- \\
ER++ & \textbf{18.45\%} & -- \\
\bottomrule
\end{tabular}
\end{table}

\subsubsection{Online large-scale classification on CLOC}
In CLOC~\citep{cai2021online}, each image in a chronological data-sequence is associated with the geographical location where it was taken, discretized to 713 (balanced) classes.  It is a highly non-stationary task 
on multiple overlapping time-scales because, e.g., major sports events lead to busts of photos from 
certain locations; seasonal changes effect the appearance of landmarks; locations become more or less popular over time; etc.
We use the version of CLOC described in \cite{bornschein2022sequential}: 
around 5\% of the images could not be downloaded or decoded which leaves us with a sequence of 37,093,769 images. 
This version of the dataset is similar to the one considered in \cite{ghunaim2023real}, but we are mindful of potential small differences due to the downloading errors. We follow the same protocol as in \cite{ghunaim2023real} and in \cite{bornschein2022sequential}.

\begin{figure}
\centering
\includegraphics[scale=0.177]{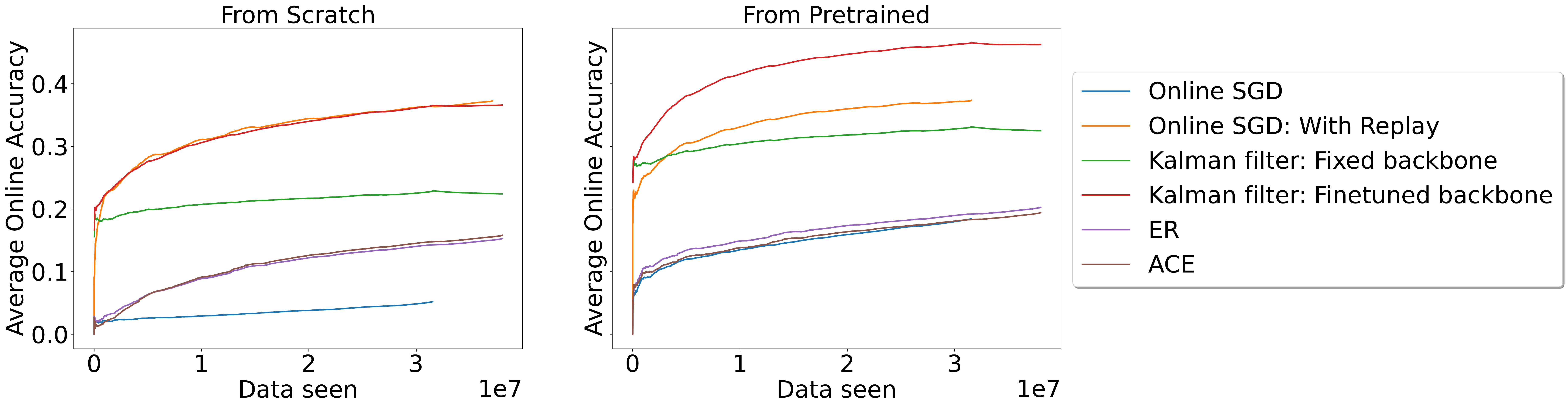}
\caption{\textbf{CLOC} results. On the left plot we present the results when we start learning from scratch. On the right plot we present results when we start from pretrained model. We also report results for external baselines taken from \citep{ghunaim2023real}: ER and ACE~\citep{caccia2022new}. Note that on the left, the top two curves that are on top of each other are Online SGD with replay and Kalman filter with finetuned backbone, while the Kalman Filter outperforms considerably on right plot. 
}
\label{fig:cloc_kalman}
\end{figure}

We use a ResNet-50 backbone and receive the data in chunks of $128$ examples. 
For the Kalman filter, we consider the variant with learned $\gamma$, which performed always better than any fixed one, including $\gamma=1$. We either keep the backbone fixed, or finetune it. 
The case of fixed backbone corresponds to a linear model only. For more details and hyperparameters selection, see Appendix~\ref{app:exp_details}. 
We run experiments where we either start learning on CLOC from scratch, or start with a ImageNet-pretrained backbone via supervised loss.
As baselines, we consider Online SGD with and without replay from \cite{bornschein2022sequential}, and
compare to the results from \cite{ghunaim2023real}: ER~\cite{chaudhry2019continual} and ACE~\citep{caccia2022new}. In order to produce the plots, we asked the authors of \cite{ghunaim2023real} to provide us the data from their experiments. The results are shown in Figure~\ref{fig:cloc_kalman}. We see that Kalman filter provides very strong performance compared to the baselines. When learning from scratch, replay-free Kalman filter matches the performance of Online SGD with replay. This is a strong result since Kalman filter does not need to store additional data in memory. Moreover, even having Kalman filter with fixed backbone performs much better than online SGD. When starting from pretrained model, Kalman filter manages to learn more efficiently than any of the baselines. Overall, this demonstrates the strong capabilities of our method to do large-scale non-stationary learning.



\section{Conclusion \label{sec_conclusion}}

We presented a probabilistic Bayesian online learning method that combines efficient Kalman filter inference with online learning of deep learning representations.  We have demonstrated 
that this method is able to adapt to non-stationarity of the data  and it can give competitive next time-step data predictions.  Some possible directions for future research are: Firstly, it would be useful to investigate  alternative approximate inference approaches for the classification case where we may construct a more accurate, but still computationally efficient, online Gaussian approximation to the softmax likelihood. Secondly, it would be interesting to extend the transition  dynamics to include higher order Markov terms, which could increase the flexibility of the algorithm to model more complex forms of non-stationarity and distribution drift. 


\bibliographystyle{plain}
\bibliography{references}

\begin{thebibliography}{10}

\bibitem{aljundi2018memory}
Rahaf Aljundi, Francesca Babiloni, Mohamed Elhoseiny, Marcus Rohrbach, and
  Tinne Tuytelaars.
\newblock Memory aware synapses: Learning what (not) to forget.
\newblock In {\em Proceedings of the European conference on computer vision
  (ECCV)}, pages 139--154, 2018.

\bibitem{aljundi2019online}
Rahaf Aljundi, Eugene Belilovsky, Tinne Tuytelaars, Laurent Charlin, Massimo
  Caccia, Min Lin, and Lucas Page-Caccia.
\newblock Online continual learning with maximal interfered retrieval.
\newblock In H.~Wallach, H.~Larochelle, A.~Beygelzimer, F.~d\textquotesingle
  Alch\'{e}-Buc, E.~Fox, and R.~Garnett, editors, {\em Advances in Neural
  Information Processing Systems}, volume~32. Curran Associates, Inc., 2019.

\bibitem{aljundi2019gradient}
Rahaf Aljundi, Min Lin, Baptiste Goujaud, and Yoshua Bengio.
\newblock Gradient based sample selection for online continual learning.
\newblock {\em Advances in neural information processing systems}, 32, 2019.

\bibitem{Blier2018-rl}
L{\'e}onard Blier and Yann Ollivier.
\newblock The description length of deep learning models.
\newblock February 2018.

\bibitem{Bommasani2021OnTO}
Rishi Bommasani, Drew~A. Hudson, Ehsan Adeli, Russ Altman, Simran Arora, Sydney
  von Arx, Michael~S. Bernstein, Jeannette Bohg, Antoine Bosselut, Emma
  Brunskill, Erik Brynjolfsson, S.~Buch, Dallas Card, Rodrigo Castellon,
  Niladri~S. Chatterji, Annie~S. Chen, Kathleen~A. Creel, Jared Davis, Dora
  Demszky, Chris Donahue, Moussa Doumbouya, Esin Durmus, Stefano Ermon, John
  Etchemendy, Kawin Ethayarajh, Li~Fei-Fei, Chelsea Finn, Trevor Gale,
  Lauren~E. Gillespie, Karan Goel, Noah~D. Goodman, Shelby Grossman, Neel Guha,
  Tatsunori Hashimoto, Peter Henderson, John Hewitt, Daniel~E. Ho, Jenny Hong,
  Kyle Hsu, Jing Huang, Thomas~F. Icard, Saahil Jain, Dan Jurafsky, Pratyusha
  Kalluri, Siddharth Karamcheti, Geoff Keeling, Fereshte Khani, O.~Khattab,
  Pang~Wei Koh, Mark~S. Krass, Ranjay Krishna, Rohith Kuditipudi, Ananya Kumar,
  Faisal Ladhak, Mina Lee, Tony Lee, Jure Leskovec, Isabelle Levent, Xiang~Lisa
  Li, Xuechen Li, Tengyu Ma, Ali Malik, Christopher~D. Manning, Suvir
  Mirchandani, Eric Mitchell, Zanele Munyikwa, Suraj Nair, Avanika Narayan,
  Deepak Narayanan, Benjamin Newman, Allen Nie, Juan~Carlos Niebles, Hamed
  Nilforoshan, J.~F. Nyarko, Giray Ogut, Laurel~J. Orr, Isabel Papadimitriou,
  Joon~Sung Park, Chris Piech, Eva Portelance, Christopher Potts, Aditi
  Raghunathan, Robert Reich, Hongyu Ren, Frieda Rong, Yusuf~H. Roohani, Camilo
  Ruiz, Jack Ryan, Christopher R'e, Dorsa Sadigh, Shiori Sagawa, Keshav
  Santhanam, Andy Shih, Krishna~Parasuram Srinivasan, Alex Tamkin, Rohan Taori,
  Armin~W. Thomas, Florian Tram{\`e}r, Rose~E. Wang, William Wang, Bohan Wu,
  Jiajun Wu, Yuhuai Wu, Sang~Michael Xie, Michihiro Yasunaga, Jiaxuan You,
  Matei~A. Zaharia, Michael Zhang, Tianyi Zhang, Xikun Zhang, Yuhui Zhang,
  Lucia Zheng, Kaitlyn Zhou, and Percy Liang.
\newblock On the opportunities and risks of foundation models.
\newblock {\em ArXiv}, abs/2108.07258, 2021.

\bibitem{bornschein2022nevis}
Jorg Bornschein, Alexandre Galashov, Ross Hemsley, Amal Rannen-Triki, Yutian
  Chen, Arslan Chaudhry, Xu~Owen He, Arthur Douillard, Massimo Caccia, Qixuang
  Feng, et~al.
\newblock Nevis'22: A stream of 100 tasks sampled from 30 years of computer
  vision research.
\newblock {\em arXiv preprint arXiv:2211.11747}, 2022.

\bibitem{bornschein2022sequential}
Jorg Bornschein, Yazhe Li, and Marcus Hutter.
\newblock Sequential learning of neural networks for prequential mdl.
\newblock In {\em The Eleventh International Conference on Learning
  Representations}, 2022.

\bibitem{buzzega2020dark}
Pietro Buzzega, Matteo Boschini, Angelo Porrello, Davide Abati, and Simone
  Calderara.
\newblock Dark experience for general continual learning: a strong, simple
  baseline.
\newblock {\em Advances in neural information processing systems},
  33:15920--15930, 2020.

\bibitem{caccia2022new}
Lucas Caccia, Rahaf Aljundi, Nader Asadi, Tinne Tuytelaars, Joelle Pineau, and
  Eugene Belilovsky.
\newblock New insights on reducing abrupt representation change in online
  continual learning, 2022.

\bibitem{cai2021online}
Zhipeng Cai, Ozan Sener, and Vladlen Koltun.
\newblock Online continual learning with natural distribution shifts: An
  empirical study with visual data, 2021.

\bibitem{chaudhry2019continual}
Arslan Chaudhry, Marcus Rohrbach, Mohamed Elhoseiny, Thalaiyasingam Ajanthan,
  Puneet~K Dokania, Philip~HS Torr, and M~Ranzato.
\newblock Continual learning with tiny episodic memories.
\newblock 2019.

\bibitem{dawid1999prequential}
A~Philip Dawid and Vladimir~G Vovk.
\newblock Prequential probability: principles and properties.
\newblock {\em Bernoulli}, pages 125--162, 1999.

\bibitem{de2021continual}
Matthias De~Lange, Rahaf Aljundi, Marc Masana, Sarah Parisot, Xu~Jia,
  Ale{\v{s}} Leonardis, Gregory Slabaugh, and Tinne Tuytelaars.
\newblock A continual learning survey: Defying forgetting in classification
  tasks.
\newblock {\em IEEE transactions on pattern analysis and machine intelligence},
  44(7):3366--3385, 2021.

\bibitem{Feldkamp1998}
Lee~A. Feldkamp, Danil~V. Prokhorov, Charles~F. Eagen, and Fumin Yuan.
\newblock {\em Enhanced Multi-Stream Kalman Filter Training for Recurrent
  Networks}, pages 29--53.
\newblock Springer US, Boston, MA, 1998.

\bibitem{FRENCH1999128}
Robert~M. French.
\newblock Catastrophic forgetting in connectionist networks.
\newblock {\em Trends in Cognitive Sciences}, 3(4):128--135, 1999.

\bibitem{ghunaim2023real}
Yasir Ghunaim, Adel Bibi, Kumail Alhamoud, Motasem Alfarra, Hasan Abed Al~Kader
  Hammoud, Ameya Prabhu, Philip~HS Torr, and Bernard Ghanem.
\newblock Real-time evaluation in online continual learning: A new paradigm.
\newblock {\em arXiv preprint arXiv:2302.01047}, 2023.

\bibitem{Grunwald2004}
Peter Grunwald.
\newblock A tutorial introduction to the minimum description length principle,
  2004.

\bibitem{HADSELL20201028}
Raia Hadsell, Dushyant Rao, Andrei~A. Rusu, and Razvan Pascanu.
\newblock Embracing change: Continual learning in deep neural networks.
\newblock {\em Trends in Cognitive Sciences}, 24(12):1028--1040, 2020.

\bibitem{hadsell2020embracing}
Raia Hadsell, Dushyant Rao, Andrei~A Rusu, and Razvan Pascanu.
\newblock Embracing change: Continual learning in deep neural networks.
\newblock {\em Trends in cognitive sciences}, 24(12):1028--1040, 2020.

\bibitem{hayes2022online}
Tyler~L Hayes and Christopher Kanan.
\newblock Online continual learning for embedded devices.
\newblock {\em arXiv preprint arXiv:2203.10681}, 2022.

\bibitem{Hazan16}
Elad Hazan.
\newblock {\em Introduction to Online Convex Optimization}.
\newblock Foundations and Trends in Optimization. Now, Boston, 2017.

\bibitem{hu2021pass}
Huiyi Hu, Ang Li, Daniele Calandriello, and Dilan Gorur.
\newblock One pass imagenet, 2021.

\bibitem{jin2020gradient}
Xisen Jin, Junyi Du, and Xiang Ren.
\newblock Gradient based memory editing for task-free continual learning.
\newblock In {\em 4th Lifelong Machine Learning Workshop at ICML 2020}, 2020.

\bibitem{kirkpatrick2017overcoming}
James Kirkpatrick, Razvan Pascanu, Neil Rabinowitz, Joel Veness, Guillaume
  Desjardins, Andrei~A Rusu, Kieran Milan, John Quan, Tiago Ramalho, Agnieszka
  Grabska-Barwinska, et~al.
\newblock Overcoming catastrophic forgetting in neural networks.
\newblock {\em Proceedings of the national academy of sciences},
  114(13):3521--3526, 2017.

\bibitem{Kozdoba19}
Mark Kozdoba, Jakub Marecek, Tigran Tchrakian, and Shie Mannor.
\newblock On-line learning of linear dynamical systems: Exponential forgetting
  in kalman filters.
\newblock In {\em Proceedings of the Thirty-Third AAAI Conference on Artificial
  Intelligence and Thirty-First Innovative Applications of Artificial
  Intelligence Conference and Ninth AAAI Symposium on Educational Advances in
  Artificial Intelligence}, AAAI'19/IAAI'19/EAAI'19. AAAI Press, 2019.

\bibitem{krizhevsky2009learning}
Alex Krizhevsky and Geoffrey Hinton.
\newblock Learning multiple layers of features from tiny images.
\newblock Technical Report~0, University of Toronto, Toronto, Ontario, 2009.

\bibitem{lee2020neural}
Soochan Lee, Junsoo Ha, Dongsu Zhang, and Gunhee Kim.
\newblock A neural dirichlet process mixture model for task-free continual
  learning, 2020.

\bibitem{li2017learning}
Zhizhong Li and Derek Hoiem.
\newblock Learning without forgetting.
\newblock {\em IEEE transactions on pattern analysis and machine intelligence},
  40(12):2935--2947, 2017.

\bibitem{lin2021clear}
Zhiqiu Lin, Jia Shi, Deepak Pathak, and Deva Ramanan.
\newblock The clear benchmark: Continual learning on real-world imagery.
\newblock In {\em Thirty-fifth Conference on Neural Information Processing
  Systems Datasets and Benchmarks Track (Round 2)}, 2021.

\bibitem{lopez2017gradient}
David Lopez-Paz and Marc'Aurelio Ranzato.
\newblock Gradient episodic memory for continual learning.
\newblock {\em Advances in neural information processing systems}, 30, 2017.

\bibitem{mai2022online}
Zheda Mai, Ruiwen Li, Jihwan Jeong, David Quispe, Hyunwoo Kim, and Scott
  Sanner.
\newblock Online continual learning in image classification: An empirical
  survey.
\newblock {\em Neurocomputing}, 469:28--51, 2022.

\bibitem{mallya2018packnet}
Arun Mallya and Svetlana Lazebnik.
\newblock Packnet: Adding multiple tasks to a single network by iterative
  pruning.
\newblock In {\em Proceedings of the IEEE conference on Computer Vision and
  Pattern Recognition}, pages 7765--7773, 2018.

\bibitem{MCCLOSKEY1989109}
Michael McCloskey and Neal~J. Cohen.
\newblock Catastrophic interference in connectionist networks: The sequential
  learning problem.
\newblock volume~24 of {\em Psychology of Learning and Motivation}, pages
  109--165. Academic Press, 1989.

\bibitem{parisi2019continual}
German~I Parisi, Ronald Kemker, Jose~L Part, Christopher Kanan, and Stefan
  Wermter.
\newblock Continual lifelong learning with neural networks: A review.
\newblock {\em Neural networks}, 113:54--71, 2019.

\bibitem{patacchiola2019deep}
Massimiliano Patacchiola, Jack Turner, Elliot~J. Crowley, Michael Boyle, and
  Amos~J. Storkey.
\newblock Bayesian meta-learning for the few-shot setting via deep kernels.
\newblock In {\em Advances in Neural Information Processing Systems}, 2020.

\bibitem{poland2005asymptotics}
Jan Poland and Marcus Hutter.
\newblock Asymptotics of discrete mdl for online prediction.
\newblock {\em IEEE Transactions on Information Theory}, 51(11):3780--3795,
  2005.

\bibitem{Puskorius94}
G.V. Puskorius and L.A. Feldkamp.
\newblock Neurocontrol of nonlinear dynamical systems with kalman filter
  trained recurrent networks.
\newblock {\em IEEE Transactions on Neural Networks}, 5(2):279--297, 1994.

\bibitem{rasmussen2006gaussian}
Carl~Edward Rasmussen and Christopher~KI Williams.
\newblock {\em Gaussian Processes for Machine Learning}.
\newblock MIT Press, 2006.

\bibitem{rusu2016progressive}
Andrei~A Rusu, Neil~C Rabinowitz, Guillaume Desjardins, Hubert Soyer, James
  Kirkpatrick, Koray Kavukcuoglu, Razvan Pascanu, and Raia Hadsell.
\newblock Progressive neural networks.
\newblock {\em arXiv preprint arXiv:1606.04671}, 2016.

\bibitem{saha2021gradient}
Gobinda Saha, Isha Garg, and Kaushik Roy.
\newblock Gradient projection memory for continual learning.
\newblock {\em arXiv preprint arXiv:2103.09762}, 2021.

\bibitem{ShalevSchwarzOL}
Shai Shalev-Shwartz.
\newblock Online learning and online convex optimization.
\newblock {\em Found. Trends Mach. Learn.}, 4(2):107–194, feb 2012.

\bibitem{shin2017continual}
Hanul Shin, Jung~Kwon Lee, Jaehong Kim, and Jiwon Kim.
\newblock Continual learning with deep generative replay.
\newblock {\em Advances in neural information processing systems}, 30, 2017.

\bibitem{LanceEKA88}
Sharad Singhal and Lance Wu.
\newblock Training multilayer perceptrons with the extended kalman algorithm.
\newblock In D.~Touretzky, editor, {\em Advances in Neural Information
  Processing Systems}, volume~1. Morgan-Kaufmann, 1988.

\bibitem{sun2022information}
Shengyang Sun, Daniele Calandriello, Huiyi Hu, Ang Li, and Michalis Titsias.
\newblock Information-theoretic online memory selection for continual learning.
\newblock {\em arXiv preprint arXiv:2204.04763}, 2022.

\bibitem{sarkka_2013}
Simo Särkkä.
\newblock {\em Bayesian Filtering and Smoothing}.
\newblock Institute of Mathematical Statistics Textbooks. Cambridge University
  Press, 2013.

\bibitem{thomee2016yfcc100m}
Bart Thomee, David~A Shamma, Gerald Friedland, Benjamin Elizalde, Karl Ni,
  Douglas Poland, Damian Borth, and Li-Jia Li.
\newblock Yfcc100m: The new data in multimedia research.
\newblock {\em Communications of the ACM}, 59(2):64--73, 2016.

\bibitem{tsiamis2020online}
Anastasios Tsiamis and George Pappas.
\newblock Online learning of the kalman filter with logarithmic regret, 2020.

\bibitem{wang2022learning}
Zifeng Wang, Zizhao Zhang, Chen-Yu Lee, Han Zhang, Ruoxi Sun, Xiaoqi Ren,
  Guolong Su, Vincent Perot, Jennifer Dy, and Tomas Pfister.
\newblock Learning to prompt for continual learning.
\newblock In {\em Proceedings of the IEEE/CVF Conference on Computer Vision and
  Pattern Recognition}, pages 139--149, 2022.

\bibitem{zenke2017continual}
Friedemann Zenke, Ben Poole, and Surya Ganguli.
\newblock Continual learning through synaptic intelligence.
\newblock In {\em International conference on machine learning}, pages
  3987--3995. PMLR, 2017.

\bibitem{zhang2020class}
Junting Zhang, Jie Zhang, Shalini Ghosh, Dawei Li, Serafettin Tasci, Larry
  Heck, Heming Zhang, and C-C~Jay Kuo.
\newblock Class-incremental learning via deep model consolidation.
\newblock In {\em Proceedings of the IEEE/CVF Winter Conference on Applications
  of Computer Vision}, pages 1131--1140, 2020.

\end{thebibliography}

\clearpage 

\appendix

\section{Connection with online Bayesian linear regression
\label{app:connectionBLR}
} 


When the forgetting coefficient is 
$\gamma_n = 1$ for all $n$, 
the stochasticity in the transitions is removed  since the parameter transition density becomes a point mass, i.e.\ $p(w_n|w_{n-1}) = \delta(w_n - w_{n-1})$.
Then, the Kalman filter 
reduces to updating the Gaussian posterior density   
$
p(w | y_{1:n}) 
= \mathcal{N}(w | m_n, A_n)
$ 
where

\begin{equation}
m_n = m_{n - 1}
+ \frac{A_{n-1} \phi_n}{\sigma^2 + \phi_n^\top A_{n-1} \phi_n}
(y_n - \phi_n^\top m_{n-1}), \ \ \  
A_n  = A_{n-1}
- \frac{A_{n-1} \phi_n \phi_n^\top A_{n-1}}{\sigma^2 + \phi_n^\top A_{n-1} \phi_n },
\label{eq:bayeslinearregression}
\end{equation}
with the initial conditions $m_0 = 0$ and $A_0 = \sigma_w^2 I$. We
see that these recursions compute 
exactly in an efficient $O(m^2)$ time the standard 
Bayesian linear regression posterior given by 
$p(w| y_{1:n}) = 
\mathcal{N}( w | A_n \sigma^{-2} \sum_{i=1}^n \phi_i y_i,  A_n)$, 
with $A_n = \left( \sigma^{-2} \sum_{i=1}^n \phi_i \phi_i^\top + \sigma_w^{-2} I \right)^{-1}$,
and where the connection with the online updates can be seen by applying the Woodbury matrix identity. Obtaining the online Bayesian linear regression recursion as a special  of a Kalman filter
is well known in the literature; see for example Section 3.2 in the book of \cite{sarkka_2013}.

\section{Online updating the forgetting coefficient $\gamma_n$ for regression
\label{app:learninggamma}
}

Learning online $\gamma_n$ for the regression case is simpler than in classification since now the 
predictive density is a tractable Gaussian given by  \eqref{eq:gaussianpredictive}. 
Thus, given that we parametrize  
$\gamma_n = \exp( - 0.5 \delta_n)$, with $\delta_n \geq 0$, the update for $\delta_n$ is written as  
\begin{equation}
\delta_n \leftarrow 
\delta_n + \rho_n \nabla_{\delta_n} \log \mathcal{N}(y_n | \phi_n^\top m_n^-, \phi_n^\top A_n^{-}  \phi_n + \sigma^2).
\label{eq:updategammaregress}
\end{equation}
The dependence of the log density on $\delta_n$ is through the parameters $(m_n^-, A_n^-)$  given by  
\eqref{eq:m_A_minus}  while the parameters $(m_{n-1}, A_{n-1})$ of the posterior up to time $n-1$ 
are taken as constants.

\section{Further details about the time series example
\label{app:toyregression}
} 
  
We generated the time series dataset consisted of $3058$ observations 
sequentially by using $8$ segments with mean values
$\{1.3, 1.0, 1.3, 0.95, 0.6, 0.25, 0.8, 0.5\}$ 
and where the change-points between these segments occurred (randomly) at the following $7$ time steps  
$\{451, 709, 958, 1547, 2147, 2769, 2957\}$. 
To obtain each observed $y_n$ we added Gaussian noise with variance $0.01$.  

For these time series data we applied a modification of the Kalman 
updates so that all updates remain the same, except for the update of the parameter $m_n^-$ which now does not shrink 
to zero and it has the form  
$$
m_n^- = m_{n-1}, \ \ \ \text{(while  before was} \ m_n^- = \gamma_n m_{n-1}). 
$$
This is appropriate in this case because shrinking the predictive posterior mean to zero by multiplying it by $\gamma_n\leq 1$ is a very strong prior assumption that $w_n$ has zero mean which does not hold, since the time series data can have arbitrary values away from zero. A more formal justification of the above is that the Markov dynamics have now the form 
$p(w_0) = \mathcal{N}(w_0|0,\sigma_w^2 I)$ and $p(w_n | w_{n-1}) = \mathcal{N}(w_n | (1-\gamma_n) \mu_{n-1} + \gamma_n w_{n-1}, \sigma_w^2 (1-\gamma_n^2) I)$ and each mean parameter $\mu_{n-1}$ when we transit from time step $n-1$ to $n$ is found by empirical Bayes so that $\mu_{n-1}=m_{n-1}$ and where 
$m_{n-1}$ is the mean of $p(w_{n-1}|y_{1:n-1})$.

The hypermarameters in the experiment were set to the following values: $\sigma_w^2 = 0.01, \sigma^2 = 0.05$ 
while $\delta_0$ was intitialized to 
$0.0$ (so that $\gamma_0=1.0$) and then updated at each step by performing SGD steps with learning rate equal to $1.0$. 

\begin{figure}
\centering
\begin{tabular}{c}
\includegraphics[scale=0.35]
{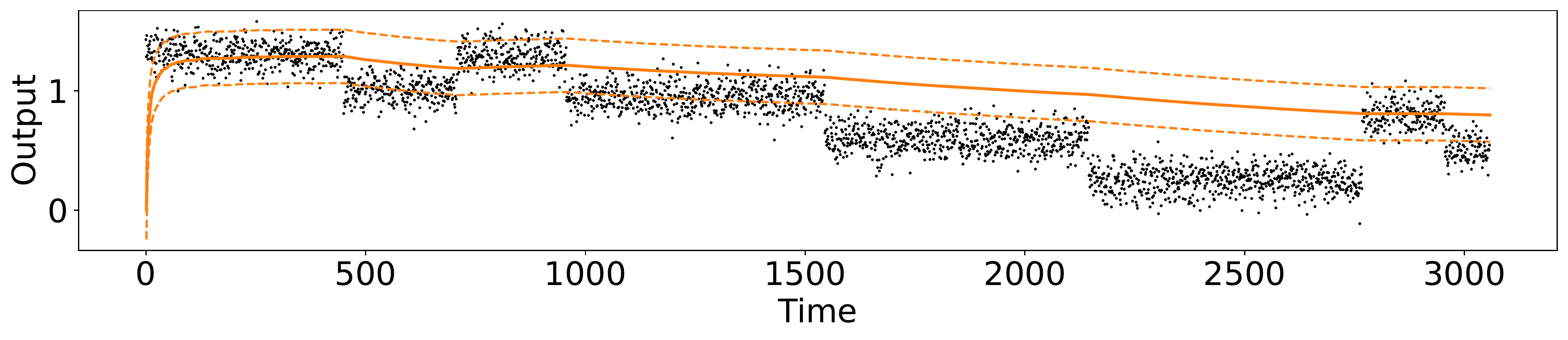} 
\end{tabular}
\caption{Online prediction on the artificial time series example by applying the Kalman filter model with fixed  $\gamma_n=1$.      
} 
\label{fig:toyregression_app}
\end{figure}

Figure \ref{fig:toyregression} in the main paper shows the results when 
$\gamma_n$ is updated online. For comparison, in  Figure \ref{fig:toyregression_app} we show  the online predictions for the case of having fixed $\gamma_n=1$ for any $n$, so that the ability to model 
non-stationarity is removed. Clearly, when  not learning $\gamma_n$ the model is not able to adjust to non-stationarity.  

\section{Experimental details and hyperparameters selection
\label{app:exp_details}
} 

\subsection{Hyperparameters selection}

For hyperparameters we do the grid search over ranges. The hyperparameters for methods are selected by choosing the variant with highest cumulative log probabilities following MDL principle from \citep{bornschein2022sequential}. When finetuning backbone together with learning calibration parameter $\alpha$, we considered a different relative scaling $\beta$ of the gradients of the backbone compared to $\alpha$ - meaning that if we use learning rate $\eta$ for $\alpha$, the learning rate for backbone is $\beta * \eta$. In case where we learned $\gamma$, we initialized $\gamma$ at $\gamma_{init}$ which is also a hyper-parameter. For each of the method, we optimized the hyper-parameters separately. The hyper-parameters and the considered values are given in the Table~\ref{table:hparams_cifar10}.

\begin{table}[!htbp]
\small
\centering
\caption{Hyper-parameters and considered values.} 
\label{table:hparams_cifar10}
\begin{tabular}{|l|l|}
\toprule
Hyperparameter & Considered values \\
\midrule
Learning rate for backbone and $\alpha$, $\eta$ & [0.1, 0.01, 0.001, 0.0005, 0.0001] \\
\midrule
Learning rate for $\gamma$, $\eta_{\gamma}$ & [0.5, 0.1, 0.01, 0.001] \\
\midrule
Relative backbone-$\alpha$ gradient scaling, $\beta$ & [0.01, 0.1, 1., 10., 100.0] \\
\midrule
Initial $\gamma_{init}$ & [0.9, 0.99, 0.999, 1.0] \\
\bottomrule
\end{tabular}
\end{table}

\subsection{Finetuning protocol}
\label{app:finetuning_protocol}

The algorithm~\ref{alg:kalman_algorithm} describes Kalman filter online learning where the learning happens for each data point. We adapt this algorithm for chunk-based learning in the following way. When we receive the chunk of data of certain size, we use the available Kalman statistics in order to calculate predictive log-probabilities (see Section~\ref{sec:finetuning}) for each data point in the chunk \textbf{independently}. After that, we do the gradient update on the backbone and $\alpha$, leading us new backbone parameters and new $\alpha$. After that, we re-compute the features on the same chunk of data and we do Kalman recursion on this chunk going through it sequentially. This latter process only involves updating statistics and $\gamma_{n}$ and doesn't involve backbone or $\alpha$ finetuning. When we do this Kalman recursion, we consider two options on how to apply \textbf{Markov transition}. In the first option, \textbf{Always Markov}, we apply Markov transition for every data point during Kalman recursion. In the second option, \textbf{Last Step Markov}, we only do Markov transition on the last point in the chunk. The difference basically lies in what we consider a non-stationary data point. In case of \textbf{Always Markov}, it is each data point in the chunk. In case of \textbf{Last Step Markov}, it is the whole chunk. We found that for different scenarios, different strategies led to different results.

\subsection{Additional per-benchmark parameters
\label{app:additionalparameters}
}

On top of the hyper-parameters which we selected for each of the baseline separately, we also found that there was a group of hyper-parameters which generally provided consistently good results for all the baselines for each benchmark. These parameters are the following:
\begin{itemize}
    \item Transition type: \textbf{Always Markov} or \textbf{Last Step Markov}
    \item Bias in features: Whether we add bias to the features or not
    \item Features normalization: Whether we normalize features vector, i.e., whether we divide it by the square root of its dimensionality.
    \item Optimization algorithm: plain SGD or Adam with weight decay ($\lambda$)
\end{itemize}

These parameters were found empirically and similarly chosen via MDL principle. The table~\ref{table:per_benchmark_parameters} gives the summary.

\begin{table}[!htbp]
\small
\centering
\caption{Per benchmark parameters.} 
\label{table:per_benchmark_parameters}
\begin{tabular}{|l|l|l|l|l|}
\toprule
Benchmark & Transition type & Use Bias & Use normalization & Optimization alg. \\
\midrule
Stationary CIFAR-100 & Last Step Markov & No & Yes & SGD \\
Non-Stationary CIFAR-100 & Always Markov & Yes & Yes & SGD \\
CLOC with pretrained backbone & Always Markov & No & No & SGD \\
CLOC from scratch & Last Step Markov & No & No & AdamW ($\lambda=10^{-4}$)  \\
\bottomrule
\end{tabular}
\end{table}

\subsection{CIFAR-100 experiments}

In both cases of CIFAR-100 experiments, we follow protocol described in \cite{ghunaim2023real}. We use ResNet-18 as backbone model and SGD as learning algorithm. We split data into chunks of size $10$. In some cases, we allow a replay strategy suggested in \cite{ghunaim2023real}, where we keep a memory of size $100$ containing most recently seen examples. Then, for each learning iteration, we sample a chunk of size $10$ from memory and append it to the right of the current chunk, creating a chunk of size $20$.

\subsection{CLOC experiments}

Similar to \citep{ghunaim2023real} and \citep{bornschein2022sequential}, we considered ResNet-50 as backbone and chunk size equal to $128$. Empirically, we found that SGD optimization 
worked much better for pre-trained model whereas Adam with weight decay ($\lambda=10^{-4}$) was working much better for learning from scratch. Moreover, we also found that when learning from scratch, it worked much better if we used Last Step Markov transition (see Appendix~~\ref{app:finetuning_protocol}). For pre-trained model, it worked better use Always Markov transition (see Appendix~~\ref{app:finetuning_protocol}).

For baselines, we considered Online SGD baseline from \cite{bornschein2022sequential}, with EMA parameter equal to 1 for fair comparison. The case of online SGD with replay corresponds to using 8 replay streams on top of the learning stream which make sure that the data distribution in the replay buffer is well behaved (see \cite{bornschein2022sequential} for more details).

\section{The effect of calibration on online classification
\label{app:calibration}
} 

\begin{figure}
\centering
\includegraphics[scale=0.2]{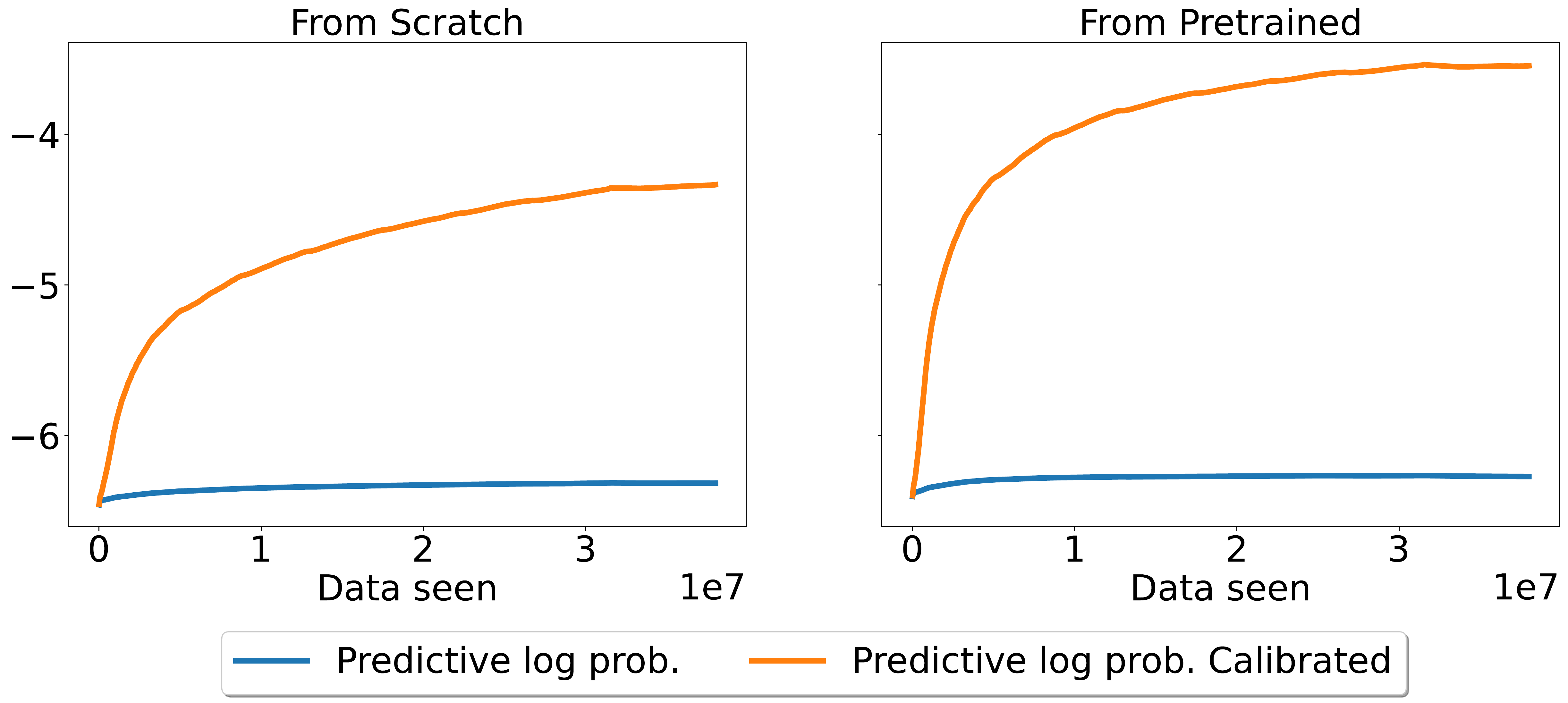}
\caption{\textbf{CLOC} log probabilitites for Kalman filter with finetuned backbone and finetuned delta. We show the data (black dots) and the predicted mean and uncertainty (orange lines) over $y_n$ (as data arrive sequentially from left to right and we perform online next time step  prediction).}
\label{fig:cloc_log_probs}
\end{figure}

\begin{figure}
\centering
\includegraphics[scale=0.2]{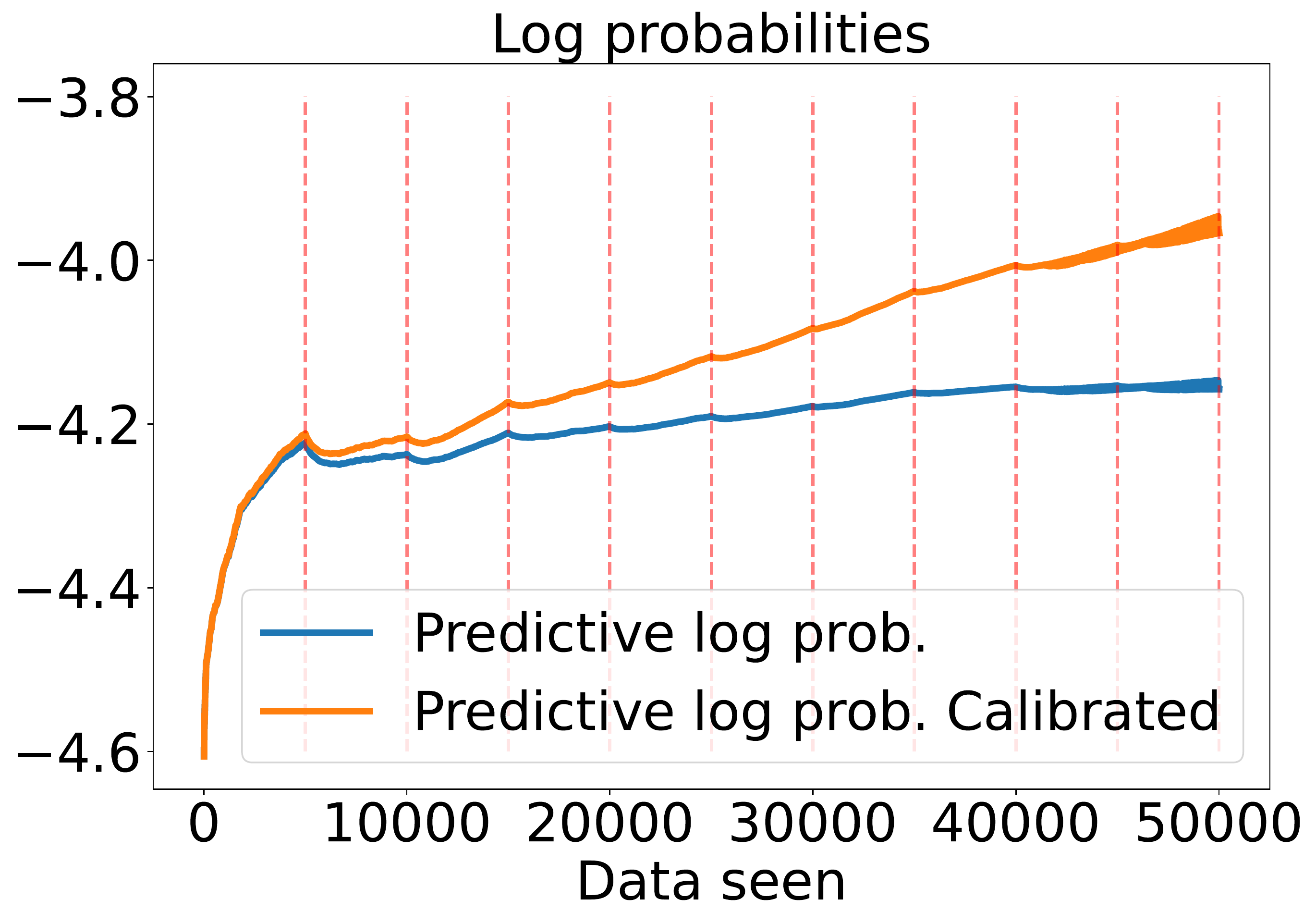}
\caption{\textbf{Non-stationary CIFAR-100} log probabilitites for Kalman filter with finetuned backbone and finetuned $\delta$.}
\label{fig:cifar_log_probs}
\end{figure}

In Section~\ref{sec:finetuning}, we describe a procedure to finetune the model backbone $\phi$ as well as to finetune the parameter $\alpha$ which affects the predictive log probability. This parameter $\alpha$ essentially allows us to calibrate the predictive log probabilities. The effect of this calibration is shown in Figure~\ref{fig:cloc_log_probs} for CLOC dataset and in Figure~\ref{fig:cifar_log_probs} for non-stationary CIFAR-100. In both cases, we use the version of Kalman filter which finetunes the backbone and learns $\gamma$. We see a very drastic positive effect of calibration -- calibrated log probabilities become much higher. Related to the discussion of prequential MDL~\citep{bornschein2022sequential}, it essentially allows to have a model with lower description length. Moreover, since we use predictive log probability as a learning signal for backbone, the calibration allows to tune the scaling of this term.

\section{Chunk size ablation
\label{app:chunk_size_ablation}
} 

\begin{figure}
\centering
\includegraphics[scale=0.2]{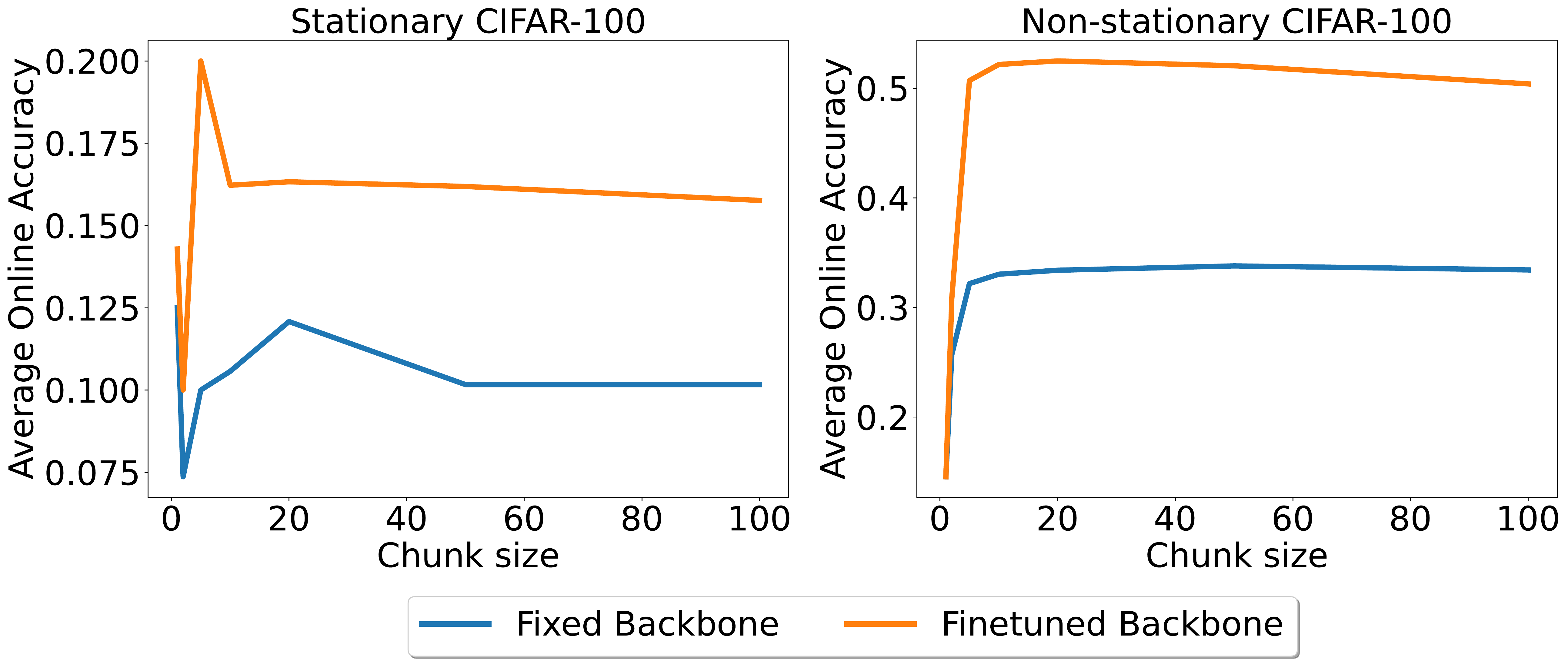}
\caption{Chunk size ablation for CIFAR-100.}
\label{fig:chunk_size_ablation_cifar100}
\end{figure}

In this section we provide the ablation of the impact of the chunk size on the average online accuracy. We provide it for CIFAR-100 as well as for CLOC. We use the Kalman filter variant with learned $\gamma$ with either fixed or finetuned backbone. The results are given in Figure~\ref{fig:chunk_size_ablation_cifar100}.

\end{document}